\documentclass[10pt,twocolumn,letterpaper]{article}

\usepackage[pagenumbers]{cvpr} 
\newtheorem{question}{Question}
\newtheorem{observation}{Observation}
\usepackage{booktabs}       %
\usepackage{tabularx}
\usepackage{multirow}

\usepackage[dvipsnames]{xcolor}

\definecolor{cvprblue}{rgb}{0.21,0.49,0.74}
\usepackage[pagebackref,breaklinks,colorlinks,citecolor=cvprblue]{hyperref}

\def\paperID{15026} 
\def\confName{CVPR}
\def\confYear{2024}

\title{Reviving Undersampling for Long-Tailed Learning}

\author{
    Hao Yu, Yingxiao Du\thanks{Equal contribution.}, Jianxin Wu\thanks{Jianxin Wu is the corresponding author.} \\
    National Key Laboratory for Novel Software Technology, Nanjing University, China\\
    {\tt\small \{yuh, duyx\}@lamda.nju.edu.cn,wujx2001@gmail.com}
}

\begin{document}
\maketitle
\begin{abstract}
The training datasets used in long-tailed recognition are extremely unbalanced, resulting in significant variation in per-class accuracy across categories.	Prior works mostly used average accuracy to evaluate their algorithms, which easily ignores those worst-performing categories. In this paper, we aim to enhance the accuracy of the worst-performing categories and utilize the harmonic mean and geometric mean to assess the model's performance. We revive the balanced undersampling idea to achieve this goal. In few-shot learning, balanced subsets are few-shot and will surely under-fit, hence it is not used in modern long-tailed learning. But, we find that it produces a more equitable distribution of accuracy across categories with much higher harmonic and geometric mean accuracy, and, but lower average accuracy. Moreover, we devise a straightforward model ensemble strategy, which does not result in any additional overhead and achieves improved harmonic and geometric mean while keeping the average accuracy almost intact when compared to state-of-the-art long-tailed learning methods. We validate the effectiveness of our approach on widely utilized benchmark datasets for long-tailed learning. Our code is at \href{https://github.com/yuhao318/BTM/}{https://github.com/yuhao318/BTM/}.
\end{abstract}    
\section{Introduction}
\label{sec:intro}
With the blessing of many balanced large-scale high-quality datasets, such as ImageNet~\cite{deng2009imagenet} and Places~\cite{zhou2018places}, deep neural networks have made significant breakthroughs in many computer vision tasks. These large-scale datasets are balanced, i.e., the number of samples in each class will be close. However, in many practical applications, the data tends to follow a long-tailed distribution, that is, the number of training images in each category is severely imbalanced. In order to solve the long-tailed recognition problem, researchers have proposed many long-tailed recognition algorithms~\cite{ren2020bsce,kang2019decoupling,he2021dive,cui2021paco,zhu2024rectify}, which achieved high average accuracy on many long-tailed datasets.

Previous long-tailed classification algorithms tend to manually split all classes into ``few'', ``medium'' and ``many'' subsets based on the number of training samples in each class, and the accuracy within each subset is usually reported along with the overall test set accuracy. However, focusing on average accuracy alone is too crude, as some worst-performing classes have zero accuracies and are overshadowed by other ``simple'' classes. Furthermore, classes in the ``few'' subset do not necessarily perform worse than those in the ``medium'' or ``many'' subsets~\cite{du2023gml}. Although the average accuracy is widely used in long-tailed classification as an optimization target, the industrial community considers the accuracies of those worst categories more critical. Therefore, it is not enough to focus on improving the average accuracy---worst-performing categories need more attention. Because the harmonic and geometric mean of per-class accuracy are more sensitive to the worst categories, GML~\cite{du2023gml} applies these metrics to measure the performance of the worst categories. Since the harmonic mean is numeric unstable to be optimized, GML chooses to maximize the geometric mean of per-class recall.

In this paper, we believe that compared to the geometric mean, the harmonic mean can better reflect the performance of the worst categories. To help the worst-performing categories, we argue that we need to \emph{revive undersampling}: using few-shot \emph{balanced} subsets to train models for long-tailed learning. Balanced undersampling has never been popular or even used practically in long-tailed learning, because it obviously will cause severe under-fitting. But, we find that on top of a regularly learning backbone network, \emph{fine-tuning on a few-shot balanced subset can (surprisingly) improve the harmonic and geometric mean greatly}, while only slightly decreasing the average accuracy. Our next surprising finding is that we can ensemble several models fine-tuned on multiple balanced few-shot datasets by directly averaging the model weights. This model averaging not only improves harmonic and geometric, but also adds no extra inference cost because its final model is a single network instead of many networks.

We name our plug-and-play and efficient training strategy as Balanced Training and Merging (BTM). In particular, our contributions are as follows:

\begin{itemize}
    \item We discover that balanced training drives the model to produce a more uniform recall distribution across categories, and averaging the fine-tuned models can further improve the harmonic and geometric mean.
    \item Based on our observations, we propose a novel plug-and-play training strategy, i.e., Balanced Training and Merging (BTM). With only a small number of samples and a little additional training overhead, BTM can significantly improve the worst-performing categories with no additional inference overhead.
    \item Our BTM is easy-to-implement, light-weight and can be integrated with other long-tailed classification algorithms easily. We conduct extensive experiments and demonstrate our effectiveness on many large-scale long-tail datasets.
\end{itemize}

\section{Related work}
\label{sec:formatting}
We review long-tailed learning methods in this section.
\subsection{Re-sampling and re-weighting methods.} Re-sampling methods either over-sample minority categories~\cite{chawla2002smote,shen2016relay,park2022context-rich-minority} or under-sample majority categories~\cite{he2009learning-from-imbalanced-data,drummond2003c4.5-class-imbalance}. There are also some methods that try to augment the minority classes by transferring the statistics obtained from the majority classes~\cite{liu2019oltr,kim2020m2m,li2021metasaug}. Re-weighting methods~\cite{lin2017focal-loss,cao2019ldam,ren2020bsce}, on the other hand, assign different weights to each category when defining the loss function. Re-sampling has the potential risk of either over-fitting or under-fitting, while re-weighting makes the loss function hard to optimize. Our BTM is based on undersampling but \emph{avoids under-fitting}. 

\subsection{Decoupling methods.} LDAM~\cite{cao2019ldam} defers re-weighting during training and cRT~\cite{kang2019decoupling} first trains a network using a plain cross-entropy loss and then re-trains the classifier using a balanced sampler in the second stage. These methods are based on the observation that over-sampling has negative effects on the learned feature representations, but is critical for learning an unbiased linear classifier~\cite{kang2019decoupling}. MiSLAS~\cite{zhong2021improving-calibration} further take model calibration into consideration and use mixup~\cite{zhang2017mixup} and label-aware smoothing~\cite{szegedy2016label-smoothing,zhong2021improving-calibration} in the first and second stage respectively. Our BTM adds an additional plug-and-play balanced training stage to the two-stage approach, but \emph{only requires little computational overhead and incurs no inference overhead}.

\subsection{Ensemble methods.} LFME~\cite{xiang2020learning-from-multiple-experts} divides the whole training set into several smaller but less imbalanced subsets and trains one model on each subset separately. Knowledge distillation~\cite{hinton2015kd} is applied later to combine these models together. BBN~\cite{zhou2020bbn} uses two branches that use different sampling strategies during training. RIDE~\cite{wang2021ride} attaches multiple heads to a single network and uses an additional loss function during training to increase the diversity of each head. During inference, special routing rules are applied to select appropriate heads for prediction. SADE~\cite{zhang2022self} aggregates the learned multiple experts for handling unknown test class distributions. Our BTM approach also merges multiple models trained on some randomly sampled few-shot datasets, but we focus on improving the accuracy of those worst-performing categories rather than the overall average accuracy.

\subsection{Other methods.} Besides the methods mentioned above, there are methods~\cite{yang2020rethinking-the-value-of-labels,wang2021contrastive-learning-based-hybrid,cui2021paco} that try to use self-supervised learning to tackle the long-tailed recognition problem. For example, PaCo~\cite{cui2021paco} uses a balanced supervised contrastive loss~\cite{khosla2020supcon}. There are methods that use knowledge distillation~\cite{he2021dive,li2021ssd} or formulate long-tailed recognition as a label-shift distribution problem~\cite{zhang2021disalign,menon2021long-tail-learning-via-logit}. There are also methods that utilize supervision signals from other modalities~\cite{long2022retrieval_augmented}. But these works usually use a lot of additional training data. Recently, Du and Wu~\cite{du2023gml} propose GML to focus more on the worst categories and propose to use the harmonic and geometric mean of per-class accuracy instead of the overall accuracy on the whole test set as an alternative metric. Our work shares the same goal as GML but achieves higher harmonic and geometric mean. Later we will also show that \emph{our BTM can be combined with GML to obtain better results}.

\section{Method} \label{sec:method}

We describe our framework in this section, starting by introducing the evaluation metrics we prefer, followed by novel questions and key observations we revealed in two-stage decoupling methods. Based on these observations, we propose our training pipeline, Balanced Training and Merging (BTM), a simple plug-and-play strategy to improve results of the worst-performing categories.

\subsection{Harmonic Mean is the Preferred Evaluation Metric}\label{sec:metric}

For long-tailed learning, given a real number $p$ and the per-class accuracy $\{ x_1, x_2, \cdots, x_n \} $ on a balanced test set, the generalized mean with exponent $p$ of these accuracies is 
\begin{equation}
  {\displaystyle M_{p}(x_{1},\dots ,x_{n})=\left({\frac {1}{n}}\sum _{i=1}^{n}x_{i}^{p}\right)^{{1}/{p}}.}
\end{equation}
For instance, when $p=-\infty$, $M_{-\infty}(x_1,\dots,x_n) = \min \{x_1,\dots,x_n\}$ is the minimum of per-class accuracy. When $p$ is $-1$ and $0$, $M_{-1}(x_1,\dots,x_n) = \frac{n}{\frac{1}{x_1}+\dots+\frac{1}{x_n}}$ and $M_{0}(x_1,\dots,x_n) = \sqrt[n]{x_1\cdot\dots\cdot x_n}$ are the harmonic and geometric mean, respectively. In particular, the arithmetic mean accuracy, $M_{1}(x_1,\dots,x_n) = \frac{x_1 + \dots + x_n}{n}$, is frequently-used in long-tailed learning. 

Compared to average accuracy, worst-case accuracy may be more important~\cite{du2023gml}. For example, given the per-class accuracy $\{x_1, x_2\} = \{0.1, 0.9\}$, its arithmetic mean is $\frac{0.1 + 0.9}{2} = 0.5$. While this result seems well, it has a minimum accuracy of 0.1, which indicates this algorithm is unusable in real-world applications. Compared to the geometric mean $\sqrt{0.1 \times 0.9} = 0.3$, the harmonic mean $\frac{2}{\frac{1}{0.1}+\frac{1}{0.9}}=0.18$ is more sensitive to the low recall values and has smaller absolute value, which is closer to $M_{-\infty}=0.1$, the worst-case accuracy we want to maximize.

However, it is hard to optimize the minimum accuracy directly.  The harmonic mean is defined using reciprocal, which makes it hard and numerically unstable to be optimized~\cite{du2023gml}. It is worth noting that every 1\% improvement in harmonic mean is very difficult and even some state-of-the-art long-tail recognition algorithms have higher average accuracy with lower harmonic mean. The previous work GML chooses to maximize the geometric mean over a mini-batch as a surrogate for the harmonic mean accuracy. In this paper, our proposed BTM applies balanced fine-tuning of the pre-trained model, which is a solution to directly improve the harmonic mean as well as the geometric mean.

\subsection{Can We Revive the Undersampling Strategy?}

\begin{figure}[t]
    \centering
   \subcaptionbox{\label{fig:hmean_ori_ft}}{\includegraphics[width=0.47\linewidth]{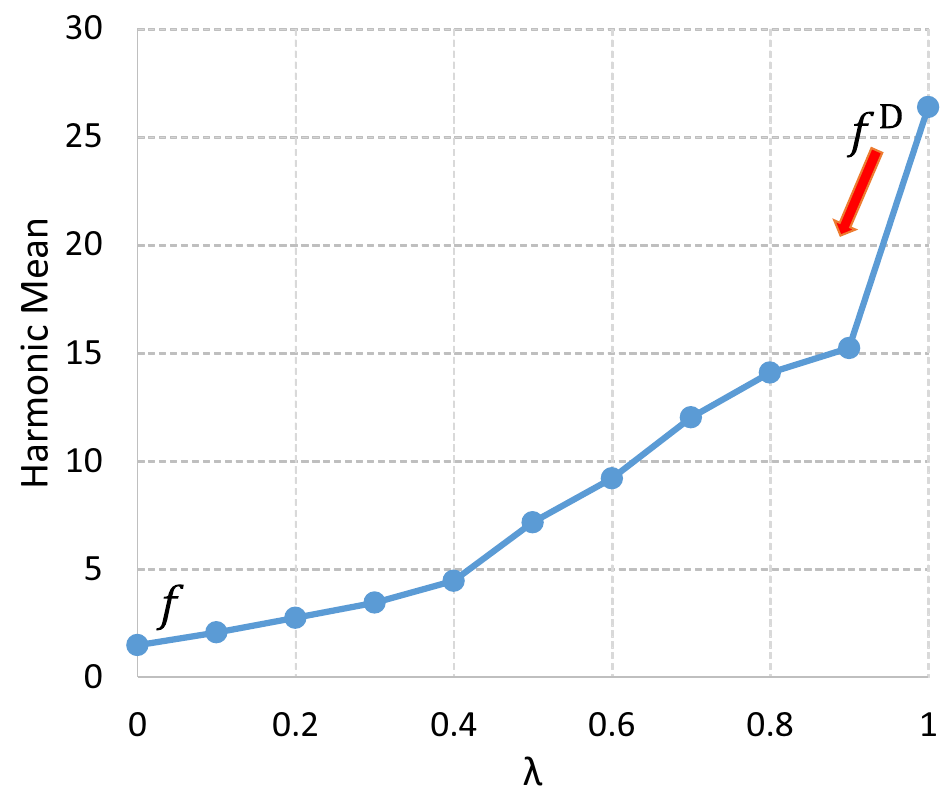}}
    \hspace{0.1cm}
    \subcaptionbox{\label{fig:gmean_ori_ft}}{\includegraphics[width=0.47\linewidth]{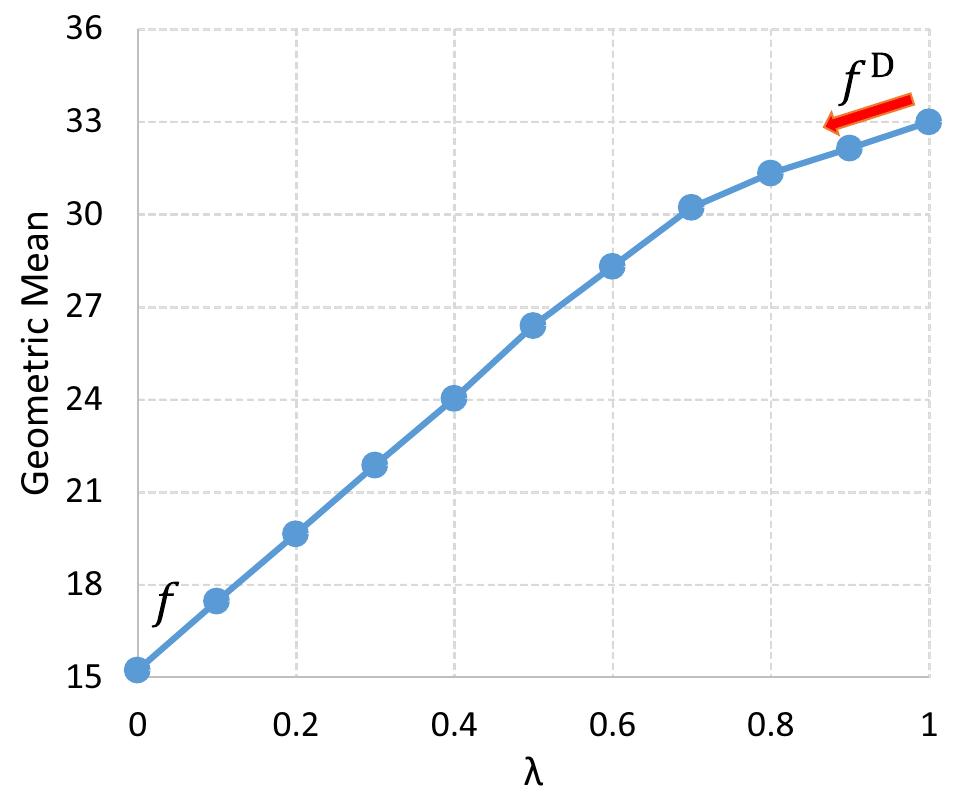}}
  \caption{\protect\subref{fig:hmean_ori_ft} and \protect\subref{fig:gmean_ori_ft} present the harmonic and geometric mean of interpolated models between the raw model $f$ ($\lambda=0$) and the fine-tuned model $f^{ D}$ ($\lambda=1$), respectively.  }
  \label{fig:total_ori_ft} 
\end{figure}
\begin{figure}[t]
    \centering
    \subcaptionbox{\label{fig:hmean_ft_ft}}{\includegraphics[width=0.47\linewidth]{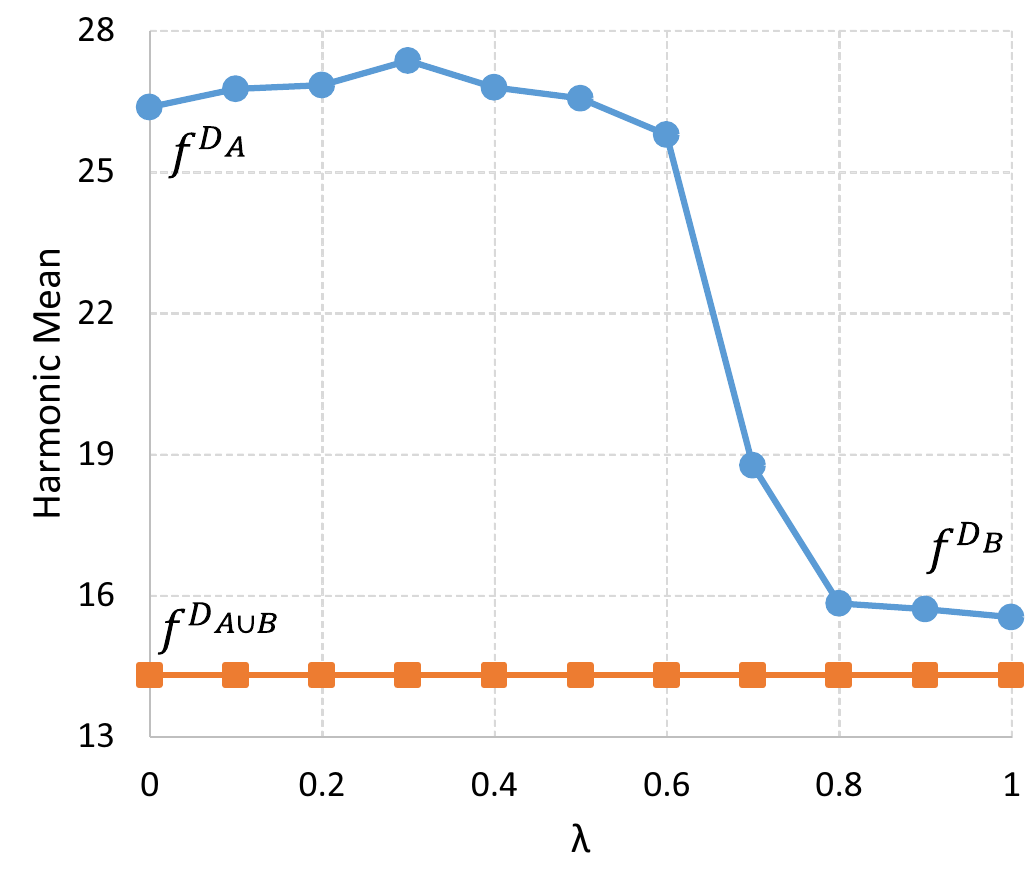}}
    \hspace{0.1cm}
    \subcaptionbox{\label{fig:gmean_ft_ft}}{\includegraphics[width=0.47\linewidth]{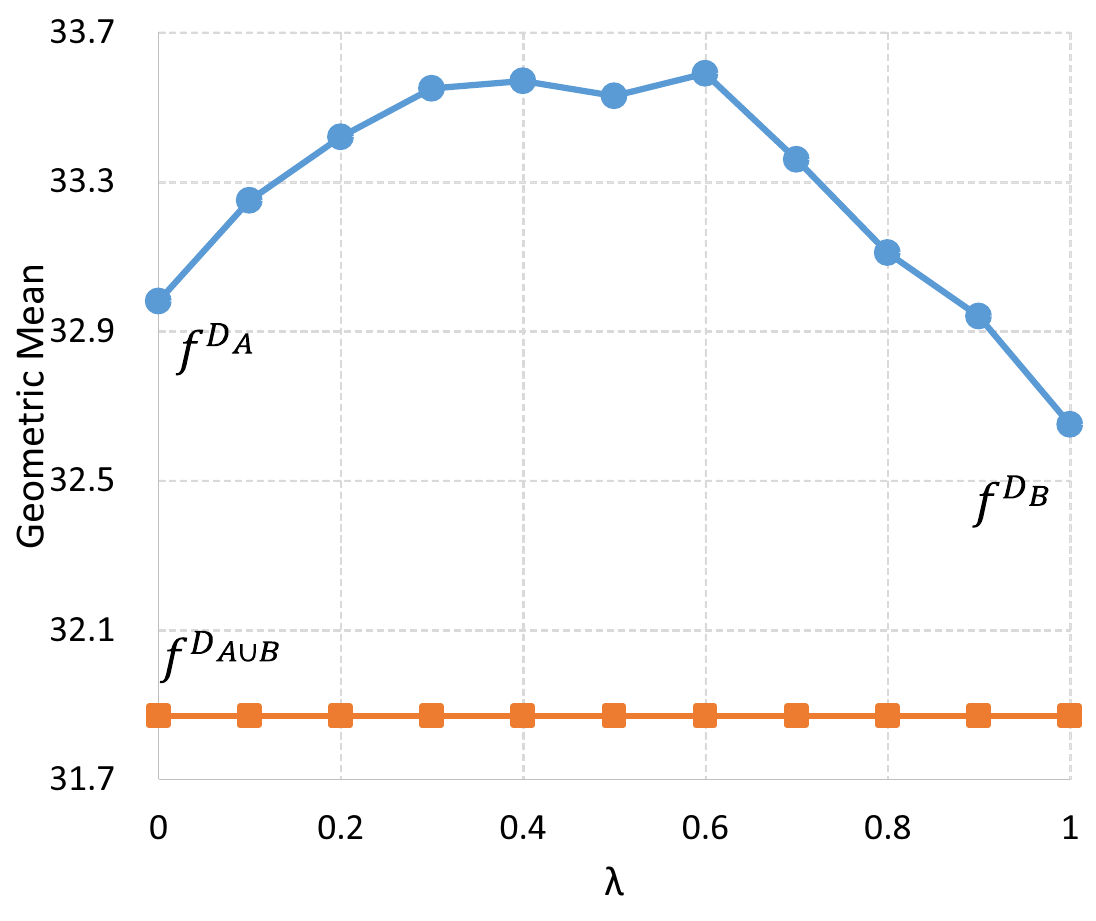}}
    \caption{The blue curves in \protect\subref{fig:hmean_ft_ft} and \protect\subref{fig:gmean_ft_ft} present the harmonic and geometric mean of interpolated models between the fine-tuned model $f^{ D_A}$ ($\lambda=0$) and the fine-tuned model $f^{ D_B}$ ($\lambda=1$), respectively. The yellow curves mean the harmonic and geometric mean of $f^{ D_{A\cup B}}$.}
    \label{fig:total_ft_ft} 
\end{figure}  

As is shown in GML, the per-class accuracy of models trained on an imbalanced dataset varies a lot from category to category. There are two recognized reasons for that. First, some categories are essentially more difficult than others. Second, there remains a large difference between the numbers of samples in different categories~\cite{he2009learning-from-imbalanced-data,chawla2002smote}. The first difficulty stems from the property of each category itself and is hard to handle. In this paper, we focus on the second difficulty and try to deal with the severely imbalanced data distribution.

To solve the difficulty induced by imbalance, \emph{the most natural solution is to have a balanced training set}. As oversampling leads to severe overfitting, undersampling seems a better choice. But, in long-tailed datasets, tail categories often have very limited (e.g., 5) training samples. Hence a \emph{balanced under-sampled} dataset will be \emph{few-shot}. A direct guess is that few-shot balanced data will inevitably cause under-fitting so as to incur catastrophic accuracy drop. Therefore, a key question is:
\begin{question}
    \label{question:few-shot_dataset}
    Can we improve the accuracy of the worst categories with the few-shot balanced dataset?
\end{question}

We conducted a simple experiment to answer this question. In particular, two-stage decoupling methods, such as cRT~\cite{kang2019decoupling} and MiSLAS~\cite{zhong2021improving-calibration}, train the whole network in the first stage, and then fine-tune the classifier in the second stage. Here we take the ResNet-50~\cite{he2016deep-residual} pre-trained with \emph{first stage} MiSLAS in the Places-LT~\cite{zhou2018places,liu2019oltr} dataset as the original model $f$, and randomly sample a 5-shot \emph{balanced} dataset $D$ from the training data. Then we fine-tune $f$ using $D$ for 30 epochs and obtain a model $f^{ D}$. After fine-tuning, following~\cite{li2020few,frankle2020linear,wang2023practical}, we \emph{merge} the original and fine-tuned models by linear interpolation. Given $\lambda\in[0, 1]$, the interpolated model is 
\begin{equation}
f_{\lambda}^{ D} =\lambda f^{ D} + (1-\lambda)f \,.
\end{equation}

Figure~\ref{fig:hmean_ori_ft} and~\ref{fig:gmean_ori_ft} show the harmonic and geometric mean of interpolated models. As we argued in the previous subsection, these curves (especially the harmonic mean curve in Figure~\ref{fig:hmean_ori_ft}) roughly describe the performance of the worst-performing categories. Surprisingly, we have one observation from the results, i.e., the harmonic and geometric mean of $f^{D}$ are $26.37\%$ and $32.98\%$, which are significantly higher than the raw model $f$ ($1.47\%$ harmonic and $15.22\%$ geometric mean). Those results prove that 
\begin{observation}
    \label{obser:finetune_improvement} For the first-stage pre-trained model in long-tailed learning, fine-tuning with a few-shot balanced dataset can highly improve the accuracy in the worst-performing categories.
\end{observation}

Besides, we also find that with the decrease of $\lambda$, the harmonic and geometric mean of the interpolated model $f_{\lambda}^{D}$ show a monotonically decreasing trend, which indicates

\begin{observation}
  The harmonic and geometric mean of these interpolated models present smooth and monotonic curves.\label{obser:finetune_interpolated}  
\end{observation}

The two surprising observations answer Question~\ref{question:few-shot_dataset}: we \emph{can} fine-tune the \emph{whole} network with only \emph{few} training samples. Even if only scarce training data is available, the fine-tuning optimization process hardly suffers from overfitting. The average accuracy of $f^D$ is 9.09\% higher than $f$, and we do obtain dramatic improvement in terms of both harmonic and geometric mean, indicating that the worst-performing categories (what we care most) have been successfully improved. That is, if we pay attention to the proper metric, we can \emph{revive} undersampling. Then, a natural question is:
\begin{question}
    \label{question:improve}
    Can we further improve both worst-case and average accuracy with few-shot balanced undersampling?
\end{question}

Note that Observations~\ref{obser:finetune_improvement} and~\ref{obser:finetune_interpolated} only consider the connection between the raw model $f$ and its fine-tuned version $f^{D}$. In particular, Observation~\ref{obser:finetune_interpolated} gives us an insight into how interpolated models might behave if we have multiple models fine-tuned on \emph{different} balanced datasets. Therefore, we fine-tuned $f$ on different 5-shot balanced datasets $D_A$ and $D_B$ to obtain fine-tuned models $f^{ D_A}$ and $f^{ D_B}$, respectively. Then we merge them by linear interpolation, too. Given $\lambda\in[0, 1]$, the interpolated model is 
\begin{equation}
  f^{ D_{ A\overset{\lambda}{\leftrightarrow} B}} =  \lambda f^{ D_A} + (1 - \lambda) f^{ D_B}.
\end{equation}

Furthermore, we also merged the balanced data $A$ and $B$ to $A \cup B$, then fine-tuned $f$ using $A \cup B$ dataset to obtain $f^{ D_{A\cup B}}$. Note that $D_{A\cup B}$ is \emph{no longer balanced}. For example, if a tail category has only 5 training examples, this category will only have 5 distinct examples in $D_{A\cup B}$, but a head category will most likely have 10 distinct examples. Figure~\ref{fig:total_ft_ft} presents the experimental results. With different balanced tiny sets, the fine-tuned models $f^{D_A}$ and $f^{D_B}$ have higher harmonic and geometric means than the original model. Surprisingly, we also have the following observation, i.e.,
\begin{observation}
  \label{obser:finetune_two_cap} The harmonic and geometric mean of the original model $f^{ D_A}$ and $f^{ D_B}$ are both higher than the ones of $f^{ D_{A\cup B}}$.
\end{observation}

This observation suggests that \emph{balancing is the key to help the worst-performing categories}. $D_{A\cup B}$, in spite of having more training data, is imbalanced and performs worse than both $D_A$ and $D_B$. And,
\begin{observation}
  \label{obser:finetune_two_interpolated} With appropriate $\lambda$ (where $\lambda=0.5$ seems good), the interpolated models $f^{ D_{ A\overset{\lambda}{\leftrightarrow} B}}$ have higher harmonic and geometric mean than both $f^{ D_A}$ and $f^{ D_B}$.  
\end{observation}
For instance, when $\lambda = 0.5$, the harmonic and geometric mean of $f^{ D_{ A\overset{\lambda}{\leftrightarrow} B}}$ are $26.56\%$ and $33.53\%$, which are higher than \emph{both} those of $f^{ D_A}$ ($26.37\%$ and $32.98\%$) and $f^{ D_B}$ ($15.54\%$ and $26.37\%$).  This finding indicates that after balanced training, we can continue to \emph{merge the fine-tuned models to achieve higher performance} in the worst-performing categories. 

\subsection{Balanced Training and Merging}\label{sec:training_pipeline}

Based on our questions, observations and findings, we propose our algorithm, Balanced Training and Merging (BTM) to revive undersampling for long-tailed learning.

Previous decoupling methods are often divided into two steps: first, train a model using the original long-tailed training set, which does \emph{not} take care of imbalance; second, freeze the backbone network and then only fine-tune the classifier of the model.

For our training pipeline, first we \emph{randomly} sample $N_D$ few-shot balanced datasets from the whole training set, and each balanced dataset contains only $N_C$ samples for each category. Note that our previous observations are all for the pre-trained model in the first stage, so the entire training process will become
\begin{itemize}
  \item \textbf{Pre-train.} Follow the original first-stage training strategy and directly train a model \emph{without} handling imbalance (e.g., only minimizing the simple cross-entropy loss).
  \item  \textbf{BTM.} Fine-tune the \emph{whole} model (including both backbone and classifier) on these $N_D$ few-shot balanced datasets, then merge the fine-tuned $N_D$ models using simple averaging.
  \item  \textbf{Post-train.} Freeze the backbone of the merged model, then follow the original second-stage training strategy and only fine-tune the classifier.
\end{itemize}

It is obvious that we just insert a plug-and-play BTM module between the first and second stage in any existing decoupling methods. Thus it can be widely applied to various decoupling methods. Note that when merging the fine-tuned models, we directly set the weights of each model as $1/N_D$. We design it in this way on purpose to make our approach as flexible and simple as possible. If there already are the pre-trained weights of the first pre-train stage, we can skip the first step. We have shown that BTM can produce a more equitable distribution of accuracy values for the pre-trained models. Later, we will further show that after the second BTM stage, stage 3 (post-train after BTM) still maintains higher harmonic and geometric mean than the original training pipeline.

Our BTM module can be plugged into long-tailed learning methods other than the decoupling ones. However, if the pre-trained model in a long-tailed algorithm has \emph{already considered and handled} the imbalance property, it is not suitable for BTM. We propose a simple modification correspondingly, which converts the `Pre-train' stage into two stages (`Pre-train' and `FC'):
\begin{itemize}
  \item \textbf{Pre-train.} Follow the original training strategy \emph{with} imbalance handling to obtain a pre-trained model.
  \item \textbf{FC.} Use the cross-entropy loss to fine-tune \emph{only the classifier} of the model on \emph{all} training data.
\end{itemize}

Note that the `BTM' and `Post-train' stages remain unchanged, which are omitted from the above list. The overhead of fine-tuning the classifier in the `FC' stage is quite small and negligible. In other words, no matter whether imbalance is handled or not in the `Pre-train' stage, the backbone (or the representation) is always useful in our BTM framework. But, the `FC' stage needs to prepare an FC that does \emph{not} handle imbalance, which is handled in the next `BTM' stage.

In summary, compared with previous long-tailed classification algorithms, BTM only adds a balance training and merging step, so our method is simple, plug-and-play and easy to deploy online. Our BTM training strategy only involves several balanced few-shot datasets, so the training overhead can be overlooked. Besides, BTM has no effect on the model structure and generates no additional inference overhead. Later we will show that in various state-of-the-art long-tail recognition algorithms, BTM can greatly increase harmonic and geometric mean while maintaining similar arithmetic mean accuracy. To the best of our knowledge, although balanced undersampling and direct weight fusion have been explored in many machine learning tasks, they have not been successfully applied in long-tailed learning task yet. And our BTM approach is the first attempt to introduce those technologies for improving the worst-performing categories.
\section{Experiments}\label{experim}

We conducted extensive experiments on three widely used benchmark datasets. First, we introduce the datasets, evaluation metrics, and implementation details. Then we compare our method with various baseline and state-of-the-art methods. Finally, we will present ablation studies. 
\begin{table*}[t]
    \centering
    \small
    \begin{tabular}{ccccc}
    \toprule
     Dataset name & \# Categories & \# Training Images & \# Test Images & Imbalance Ratio\\
     \midrule
     ImageNet-LT~\cite{deng2009imagenet,liu2019oltr}&1,000&115,846&50,000&256\\
     Places-LT~\cite{zhou2018places,liu2019oltr}&365&62,500&36,500&996\\
     iNaturalist2018~\cite{cui2018inat}&8,142&437,513&24,426&500\\
    \bottomrule
    \end{tabular}
    \caption{Some statistics of the benchmark datasets used.}
    \label{tab:dataset}
\end{table*}

\subsection{Datasets, Metrics, and Implementation Details}

We use three widely used long-tailed recognition datasets: Places-LT~\cite{zhou2018places,liu2019oltr}, ImageNet-LT~\cite{deng2009imagenet,liu2019oltr} and iNaturalist2018~\cite{cui2018inat}. Statistics about them can be found in Table~\ref{tab:dataset}. The original Places~\cite{zhou2018places} and ImageNet~\cite{deng2009imagenet} are balanced datasets. We follow previous work~\cite{liu2019oltr} to construct the long-tailed version by down-sampling the original training set using a Pareto distribution. 
The imbalance ratio of a dataset is defined as the number of training images of the most frequent class divided by the number of images of the least frequent class. ImageNet and Places are also two larger-scale balanced datasets, and ImageNet-LT and Places-LT are their long-tailed versions built by Liu et al.~\cite{liu2019oltr}. The Pareto distribution with a power value of $alpha = 6$ is used to determine the number of training images for each class. Their original test sets are not altered.

Following GML~\cite{du2023gml}, we focus on improving the worst-performing categories in long-tailed recognition. Besides the conventional average accuracy, we compute the accuracy for each category and report their harmonic and geometric mean. These two metrics are more sensitive to small numbers than conventional accuracy, which are believed to better reflect the fairness of a model. 

Following previous work~\cite{liu2019oltr,zhong2021improving-calibration}, we use ResNet-152~\cite{he2016deep-residual} on Places-LT, ResNeXt-50~\cite{xie2017aggregated-residual} or ResNet-50 on ImageNet-LT and ResNet-50 on iNaturalist2018. We choose to apply our method to PaCo/GPaCo~\cite{cui2021paco,cui2022gpaco} and MiSLAS~\cite{zhong2021improving-calibration}, which are two current state-of-the-art long-tail recognition methods. 

For MiSLAS, we directly use the pre-train weights of the first `Pre-train' stage. Then we apply our BTM and `Post-train' stage. We first train our model 30 epochs in the BTM stage. After that, we follow the original second-stage training strategy in the MiSLAS and fine-tune the classifier 10 epochs in the `Post-train' stage. For GPaCo on Places-LT and PaCo on ImageNet-LT, we first re-train the classifier for 10 epochs with all training data in the `FC' stage. Then we fine-tune the whole model for 30 epochs with the balanced dataset in the BTM stage. In the final `Post-train' stage, we train the classifier for 40 epochs. It can be seen that in the BTM stage, we use a very small amount of data to train the entire model with only 30 epochs. Besdies, in the `FC' or `Post-train' stages, we only need to fine-tune the classification layer, so our lightweight framework only requires few computational resources. More details about the training hyper-parameter settings can be found in the appendix.


\subsection{Comparison with Other Methods}

Now we present the comparison of our methods with various baseline and state-of-the-art methods. In particular, for categories that have zero accuracy, we substitute it with a small number ($10^{-3}$) otherwise the harmonic and geometric mean will be zero. In these tables, ``H-Mean'' stands for harmonic mean, ``G-Mean'' stands for geometric mean and ``L-Recall'' stands for the lowest recall across all categories. Note that we do not report the lowest recall in both ImageNet-LT and iNaturalist2018 datasets, because their lowest recall is zero across all algorithms.

\textbf{Places-LT.}
Table~\ref{tab:Places-LT-res} shows the comparison results on Places-LT. We apply our method to GPaCo and MiSLAS on this dataset. GPaCo is an extension of PaCo, which simplifies some training settings and achieves better results. Our BTM method improves harmonic and geometric mean by large margins while maintaining the overall accuracy. For example, compared with MiSLAS, GPaCo has a higher accuracy rate and lower harmonic and geometric mean. Applying our method on GPaCo boosts the harmonic mean from 10.93\% to 29.36\% (+18.43\%) while accuracy is slightly reduced by less than 1\%. For MiSLAS, the conventional accuracy even increases along with harmonic and geometric mean. We also list the target objective (worst category's accuracy) in the `L-Recall' column, where BTM shows clear advantages, too.

\begin{table}[t]
    \centering
	\small
    \begin{tabularx}{\columnwidth}{Xcccc}
        \toprule
        Methods&H-Mean&G-Mean&L-Recall&Acc.\\
        \midrule
        CE&\phantom{0}0.73&12.11&0.00&28.71\\
        BSCE~\cite{ren2020bsce}&\phantom{0}5.64&29.30&0.00&37.18\\
        PaCo~\cite{cui2021paco}&\phantom{0}2.53&27.88&0.00&40.45\\
        MiSLAS~\cite{zhong2021improving-calibration}&28.75&35.30&3.00&40.12\\
        GPaCo~\cite{cui2022gpaco}&10.93&35.00&0.00&\textbf{41.68}\\
        \midrule
        MiSLAS + BTM&\textbf{29.80}&35.60&\textbf{4.00}&40.25\\
        GPaCo + BTM&29.36&\textbf{35.86}&2.00&40.72\\
        \bottomrule
    \end{tabularx}
    \caption{Results on the Places-LT dataset. }
    \label{tab:Places-LT-res}
\end{table}
\begin{table}[t]
    \centering
	\small
    \begin{tabularx}{\columnwidth}{Xccc}
        \toprule
        Methods&H-Mean&G-Mean&Accuracy\\
        \midrule
        CE&\phantom{0}1.25&23.25&43.90\\
        BSCE~\cite{ren2020bsce}&13.74&42.32&50.48\\
        cRT~\cite{kang2019decoupling}&13.82&41.35&49.64\\
        DiVE~\cite{he2021dive}&12.76&45.49&53.63\\
        RIDE~\cite{wang2021ride}&17.32&47.56&55.69\\
        PaCo~\cite{cui2021paco}&21.75&51.29&\textbf{58.32}\\
        \midrule
        PaCo + BTM&\textbf{22.85}&\textbf{51.45}&58.15\\
        \midrule
        ResNet-50 as backbone\\
        \midrule
        MiSLAS~\cite{zhong2021improving-calibration}&17.68&45.78&\textbf{52.68}\\
        MiSLAS + BTM&\textbf{21.11}&\textbf{45.85}&52.59\\
        \bottomrule
    \end{tabularx}
    \caption{Results on the ImageNet-LT dataset. Note that MiSLAS uses ResNet-50 instead of ResNeXt-50 as the backbone network, so we list them in separate rows.}
    \label{tab:ImageNet-LT-res}
\end{table}

\begin{table}[t]
    \centering
	\small
    \begin{tabularx}{\columnwidth}{Xccc}
        \toprule
        Methods&H-Mean&G-Mean&Accuracy\\
        \midrule
        BSCE~\cite{ren2020bsce}&1.50&43.92&67.65\\
        BBN~\cite{zhou2020bbn}&1.52&45.43&69.71\\
        DiVE~\cite{he2021dive}&1.85&49.56&71.10\\
        MiSLAS~\cite{zhong2021improving-calibration}&2.03&51.27&\textbf{71.57}\\
        \midrule
        MiSLAS + BTM&\textbf{2.18}&\textbf{52.02}&71.43\\
        \bottomrule
    \end{tabularx}
    \caption{Results on the iNaturalist2018 dataset. }
    \label{tab:iNat18-res}
\end{table}

\textbf{ImageNet-LT.}
Table~\ref{tab:ImageNet-LT-res} shows the results on ImageNet-LT. When adding our BTM to PaCo and MiSLAS, we improved the harmonic and geometric mean while the overall accuracy remained almost unchanged. In particular, we improved the harmonic mean (+1.10\% and +3.43\%) more than the geometric mean (+0.16\% and +0.07\%). For MiSLAS, since it uses ResNet-50 instead of ResNeXt-50 as the backbone, the number cannot be directly comparable to other methods. Generally speaking, BTM successfully improves the worst-performing categories and does no harm to the overall accuracy. 

\begin{table*}[t]
    \centering
	\small
    \begin{tabular}{c|c|cccc}
        \toprule
        \multicolumn{2}{c|}{Methods} &Harmonic Mean&Geometric Mean&Lowest Recall &Accuracy\\
        \hline
        \multirow{2}{*}{Original Model} & Stage1 &1.47&15.22&0.00&29.62\\
                                & Stage2 &28.75&35.30&3.00&40.12\\
        \hline
        \multirow{10}{*}{Balanced Training} & Model1  &23.85&32.92&1.00&38.70\\
                                            & Model2  &24.99&33.00&2.00&38.58\\
                                            & Model3  &25.17&33.26&1.00&38.69\\
                                            & Model4  &25.49&33.14&2.00&38.92\\
                                            & Model5  &24.87&32.93&1.00&38.55\\
                                            & Model6  &15.31&33.04&1.00&38.89\\
                                            & Model7  &24.69&32.64&2.00&38.42\\
                                            & Model8  &24.04&32.91&1.00&38.74\\
                                            & Model9  &24.54&33.09&1.00&38.84\\
                                            & Model10  &25.79&33.30&2.00&38.95\\
        \hline
        \multicolumn{2}{c|}{Merge} &26.34&34.26 &2.00&39.84\\
        \hline
        \multicolumn{2}{c|}{Post-train} &\textbf{29.80}&\textbf{35.60} &\textbf{4.00}&\textbf{40.25}\\
        \bottomrule
    \end{tabular}
    \caption{Results of single fine-tuned and merged models in Places-LT with MiSLAS. }
    \label{tab:single-ft-merge-places-mislas}
\end{table*}

\begin{table*}[t]
    \centering
	\small
    \begin{tabular}{c|c|cccc}
        \toprule
        \multicolumn{2}{c|}{Methods} & Harmonic Mean&Geometric Mean&Lowest Recall &Accuracy\\
        \hline
        \multirow{2}{*}{Average Merging} & Merge &26.34 & 34.26 & 2.00 & 39.84 \\
                                & Post-train &\textbf{29.80} & \textbf{35.60} & 4.00 & 40.25\\
        \hline
        \multirow{2}{*}{Adaptive Ratio Based on H-Mean} & Merge &26.70&34.35&2.00&39.89\\
                                & Post-train &29.73&35.53&4.00&40.27\\
        \hline
        \multirow{2}{*}{Adaptive Ratio Based on G-Mean} & Merge &26.34&34.27&2.00&39.85\\
                                & Post-train &29.77&35.57&4.00&\textbf{40.46}\\
        \hline
        \multirow{2}{*}{Greedy Soup~\cite{wortsman2022model} Based on H-Mean} & Merge &26.84 &34.32&2.00&39.81\\
                                & Post-train &29.50&35.40&4.00&40.19\\
        \hline
        \multirow{2}{*}{Greedy Soup~\cite{wortsman2022model} Based on G-Mean} & Merge &26.79&34.37&2.00&39.88\\
                                & Post-train &29.68&35.59&4.00&40.32\\

                                \bottomrule
    \end{tabular}
    \caption{Results of different weight merging strategies. }
    \label{tab:ensemble-strategy}
\end{table*}
\begin{table}[t]
    \centering
	\small
    \begin{tabular}{cccccc}
        \toprule
        $N_{\mathcal{D}}$&$N_{\mathcal{C}}$&H-Mean&G-Mean&L-Recall&Accuracy\\

        \midrule
        2&10&28.43&34.95&2.00&39.84\\
        4&10&29.11&35.12&3.00&39.96\\
        8&10&28.20&34.83&2.00&39.87\\
        20&10&15.17&32.55&0.00&38.31\\
        10&5&28.48&35.13&2.00&40.06\\
        10&10&\textbf{29.80}&\textbf{35.60}&4.00&\textbf{40.25}\\
        10&20&29.59&35.30&\textbf{5.00}&40.09\\
        \bottomrule
    \end{tabular}
    \caption{Effects of the size of the sampled datasets. $N_\mathcal{D}$ stands for the number of few-shot datasets sampled and $N_{\mathcal{C}}$ stands for the number of training images for each category. }
    \label{tab:abaltion-dataset-size}
\end{table}

\textbf{iNaturalist2018.}
Table~\ref{tab:iNat18-res} summarizes the results of the experiments conducted on the iNaturalist2018 dataset. The average accuracy with applying BTM remains almost unchanged. Compared to ImageNet-LT and Places-LT, iNaturalist2018 has a much larger scale. Furthermore, since each category only has three test images, all current methods have a very low harmonic mean of recall on this dataset. Therefore, it is difficult to improve harmonic and geometric mean on the iNaturalist2018 dataset. Nevertheless, we also obtain 0.15\% and 0.75\% improvements, respectively. 

\subsection{Ablation Studies}

We conducted several ablation studies on Places-LT with our BTM method applied to MiSLAS. If not otherwise specified, we used the default training setting. More detailed results of the single fine-tuned models and the merged models are in the supplementary material.

\begin{table*}[t]
    \centering
	\small
    \begin{tabular}{ccccccc}
        \toprule
        When&Backbone&Classifier&Harmonic Mean&Geometric Mean&Lowest Recall&Accuracy\\
        \midrule
        Between Stage1\&2&\checkmark&&29.04&35.19&4.00&40.17\\
        Between Stage1\&2&&\checkmark&29.41&35.32&4.00&40.12\\
        Between Stage1\&2&\checkmark&\checkmark&\textbf{29.80}&\textbf{35.60}&4.00&\textbf{40.25}\\
        After Stage2&\checkmark&&28.95&35.40&2.00&40.19\\
        After Stage2&&\checkmark&28.51&35.25&2.00&40.19\\
        After Stage2&\checkmark&\checkmark&28.49&35.25&2.00&40.19\\
        \bottomrule
    \end{tabular}
    \caption{When and how to perform the balanced training. }
    \label{tab:abaltion-finetuning-choices}
\end{table*}
\begin{table*}[t]
    \centering
	\small
    \begin{tabular}{ccccc}
        \toprule
        Methods&Harmonic Mean&Geometric Mean&Lowest Recall &Accuracy\\
        \midrule
        MisLas&28.75&35.30&3.00&40.12\\
        MisLas + GML&28.81&34.94&3.00&39.68\\
        \midrule
        MisLas + BTM&29.80&\textbf{35.60}&4.00&\textbf{40.25}\\
        MisLas + BTM + GML&\textbf{29.90}&35.39&4.00&39.92\\
        \bottomrule
    \end{tabular}
    \caption{Combining GML with our method. }
    \label{tab:ablation-combine-GML}
\end{table*}

\textbf{Results of Single Fine-tuned Models and the Merged Models.} In this section, we also report the results of each balanced fine-tuned and merged model during training. In particular, MiSLAS is a two-stage decoupling method. we directly use the first-stage pre-training model and balanced fine-tune ten models based on it. After the balanced-training stage, we merge those ten models and fine-tune the classifier. The results on the Places-LT dataset are in Table~\ref{tab:single-ft-merge-places-mislas}. It can be seen that compared with the first-stage pre-trained model, each model of balanced training has higher harmonic and geometric mean, and the merging strategy has further improved the results. After post-training, the accuracy of our final model produces a more even distribution of accuracy than the original model. 

\textbf{Results of Different Weight Merging Strategies.} 
In the default settings, we set the merging ratio for each model weight to $1/N_D$. In addition to the ``Average Merging'' strategy, we also explore adaptive fusion ratio strategies. In particular, the fusion coefficients are proportional to each model's harmonic and geometric mean in the balanced training dataset, and the sum of these coefficients is one. We call these strategies ``Adaptive Ratio Based on H\&G-Mean''. In addition, we also follow the averaging weights strategy of greedy soups~\cite{wortsman2022model} and use the harmonic and geometric mean as the criterion. That is, we use one model only when merging the model is better than not merging. We call these methods ``Greedy Soup Based on H\&G-Mean''. Table~\ref{tab:ensemble-strategy} shows the results of merging and post-training with different strategies. It can be seen that after merging, these strategies of adaptively adjusting the ratios and models can achieve higher harmonic and geometric mean than direct average merging. But after post-training, these temporary small advantages are quickly wiped out. The simplest average merging strategy achieves the highest harmonic and geometric mean instead. Therefore, BTM directly uses the average merging strategy for flexibility and simplicity.

\textbf{Effects of the Size of the Sampled Few-Shot Datasets.} In the default training settings, we randomly sample 10 few-shot datasets to perform the balanced training. And for each dataset, all categories have 10 training images so the sampled dataset is balanced. In this subsection, we study the effects of the size of the sampled few-shot datasets by varying the number of datasets sampled and the number of training images for each category. The results are shown in Table~\ref{tab:abaltion-dataset-size}. As we can see from the table, when we fix $N_\mathcal{C}=10$, the performance can be improved at the beginning when we increase $N_{\mathcal{D}}$ but later drops significantly when we set $N_{\mathcal{D}}=20$. One possible reason for this phenomenon is that the model is over-fitting because we use the same tail-class examples too many times (e.g., in all sampled few-shot datasets). On the other hand, when we fix $N_{\mathcal{D}}=10$ and vary $N_{\mathcal{C}}$, the performance does not change much. Generally speaking, $N_{\mathcal{D}}=10$ and $N_{\mathcal{C}}=10$ seem to be a good choice in the Places-LT dataset. For simplicity, we follow this setting across all experiments.

\textbf{When and How to Perform the Balanced Training.}
As is mentioned in the previous section, currently we add the balanced training between the first and second stages of decoupled two-stage methods and we fine-tune the whole model using our sampled few-shot datasets. Here we explore some other possible design choices. Specifically, we try to only fine-tune the backbone or classifier or add the balanced training after the second stage. The results are shown in Table~\ref{tab:abaltion-finetuning-choices}. As we can see from the table, fine-tuning either the backbone or classifier can both improve the harmonic and geometric mean of per-class accuracy, but the final results are inferior to fine-tuning the whole model. Since the scale of our sampled datasets are small, fine-tuning the whole model would not cause much training overhead, we choose to fine-tune the whole model in order to achieve better performance. As for when to perform the balanced training, from the table we can see that adding the balanced training after the second stage achieves inferior performance compared to adding it between the first and second stage.

\textbf{Combining BTM with GML.}
GML~\cite{du2023gml} is the pioneering work in long-tailed recognition that aims at improving the performance of the worst categories. Since their method is also a plug-in, here we try to combine our method with GML. Specifically, in the third stage of our method, we use GML to fine-tune the classifier. This experiment was conducted on Places-LT and the results are shown in Table~\ref{tab:ablation-combine-GML}. As we can see from this table, although GML can improve the harmonic mean, there is a noticeable drop in accuracy. BTM, on the other hand, does little harm to the overall accuracy. Furthermore, compared to using GML alone, combing our method with GML can further improve the harmonic mean, and our BTM even achieves a higher geometric mean. This proves that our proposed balanced training is indeed very helpful in improving the performance of the worst categories, and we can further combine our BTM with pioneering work to obtain better performances.

\section{Conclusions, Limitations and Future Work}

In this paper, we present a straightforward plug-and-play training strategy to tackle the worst-category problem in long-tailed learning. Based on our two surprisingly findings, by reviving (few-shot) balanced undersampling, our proposed training strategy, named BTM, can be easily integrated with various long-tailed algorithms, requiring minimal training overhead and imposing no additional inference burden. Across multiple widely used long-tailed datasets, BTM consistently achieves notable and stable improvements in both harmonic and geometric mean accuracy, while maintaining comparable average accuracy.  

Although our method can significantly improve the accuracy balance across categories, we observed that for some large long-tailed datasets such as ImageNet and iNaturalist2018, the minimum recall remains zero even with the help of BTM. As a result, an intriguing direction for future research is how can we further enhance the minimum recall. Additionally, though our BTM will substantially increase harmonic and geometric mean, it will slightly decrease arithmetic accuracy in some scenarios. Therefore, another interesting direction to explore is the simultaneous improvement of average accuracy alongside harmonic and geometric mean. 


{
    \small
    \bibliographystyle{ieeenat_fullname}
    \bibliography{main}
}

\clearpage
\setcounter{page}{1}
\maketitlesupplementary
\appendix

In this supplementary material, we provide further details of experiments conducted in the main paper together with hyperparameters for training. We then report additional ablations investigating the results of combining BTM with GML. We also show detailed results of single fine-tuned models and the merged models. Finally, we show the visualization of the per-class accuracy.


\section{Implementation Details of the Fine-tuning Stage}

\textbf{MiSLAS.}  For the fine-tuning process in the BTM stages, we follow the data augmentation strategy of the `Pre-train' stage, i.e., we resize the image by setting the shorter side to 256 and then take a random crop of $224\times 224$ from it or its horizontal flip. Besides, color jittering and mixup~\cite{zhang2017mixup} are also applied. We set the $\alpha$ of mixup as 0.2 and batch size as 256. We use stochastic gradient descent (SGD) optimizer and set the momentum and weight decay as 0.9 and $5\times 10^{-4}$. The initial learning rate is $5\times 10^{-3}$, $1\times 10^{-3}$ and $5\times 10^{-4}$ for Places-LT, ImageNet-LT and iNaturalist2018, respectively. The cosine learning rate schedule and traditional cross-entropy loss are used. 

\textbf{PaCo/GPaCo.} For PaCo with ImageNet-LT dataset, we use the same training strategy for the two steps for simplicity. The random resized crop, color jitter and mixup are used. We set the $\alpha$ of mixup as 0.2. We use stochastic gradient descent (SGD) optimizer and set the momentum and weight decay as 0.9 and $5\times 10^{-4}$. The cosine learning rate schedule and traditional cross-entropy loss are also used. In the FC stage, the initial learning rate is 0.1. And then we set the initial learning rate as $1\times 10^{-5}$ in the BTM stage. In the final `Post-train' stage, we apply the re-weighting and re-sampling training strategy. The loss function is label-aware smoothing loss and we train the classifier 40 epochs with a cosine learning rate schedule. For GPaCo on Places-LT, the training setting is the same as the first stage training of MiSLAS on Places-LT. For the BTM  stage, we change the initial learning rate to $1\times 10^{-3}$ and everything else stays the same as the previous stage.  In the final `Post-train' stage, the training setting is the same as the second stage of MiSLAS on Places-LT. 

\section{Combining BTM with GML in PaCo}
We also use GML in ImageNet-LT. This experiment is conducted on PaCo and the results are shown in Table~\ref{tab:supp-combine-GML-PACO}. As we can see from this table, although GML can improve the harmonic and geometric mean, there is a noticeable drop in accuracy. Our method, on the other hand, does little harm to the overall accuracy. Furthermore, compared to using GML alone, combing our method with GML can further improve the harmonic and geometric mean, as well as the overall accuracy. 

\begin{table}[t]
    \centering
    \begin{tabularx}{\columnwidth}{Xccc}
        \toprule
        Methods&H-Mean&G-Mean&Accuracy\\
        \midrule
        PaCo&21.75&51.29&\textbf{58.32}\\
        PaCo + GML&31.06&50.77&55.61\\
        \midrule
        PaCo + BTM&22.85&51.45&58.15\\
        PaCo + BTM + GML&\textbf{31.26}&\textbf{51.44}&56.33\\
        \bottomrule
    \end{tabularx}
    \caption{Combing GML with our method. The experiment is conducted on ImageNet-LT with our method applied to PaCo~\cite{cui2021paco}.}
    \label{tab:supp-combine-GML-PACO}
\end{table}
\section{Results of Single Fine-tuned Models and the Merged Models}
In this section, we report the detailed results of each balanced fine-tuned and merged model during training. The results on the ImageNet-LT and iNaturalist2018 datasets with MiSLAS are in the Table~\ref{tab:supp-single-ft-merge-imagenet-mislas} and Table~\ref{tab:supp-single-ft-merge-ina2018-mislas} respectively. PaCo and GPaCo are not decoupling methods and we first apply the `FC' stage and then balanced train the weights. The results on the ImageNet-LT and Places-LT datasets are in Table~\ref{tab:supp-single-ft-merge-imagenet-paco} and Table~\ref{tab:supp-single-ft-merge-placeslt-gpaco} respectively. We can still come to similar conclusions.

\begin{table*}[t]
    \centering
	\small
    \begin{tabular}{c|c|ccc}
        \toprule
        \multicolumn{2}{c|}{Methods} &Harmonic Mean&Geometric Mean &Accuracy\\
        \hline
        \multirow{2}{*}{Original Model} & Stage1 &0.93&21.28&45.51\\
                                & Stage2 &17.68&45.78&\textbf{52.68}\\
        \hline
        \multirow{10}{*}{Balanced Training} & Model1  &2.51&36.53&50.37\\
                                            & Model2  &2.70&36.28&50.33\\
                                            & Model3  &2.70&36.30&50.23\\
                                            & Model4  &2.77&36.24&50.05\\
                                            & Model5  &2.63&36.25&50.29\\
                                            & Model6  &2.78&36.60&50.48\\
                                            & Model7  &2.70&36.38&50.27\\
                                            & Model8  &2.76&36.07&50.16\\
                                            & Model9  &2.63&35.89&49.98\\
                                            & Model10  &2.63&35.98&49.97\\
        \hline
        \multicolumn{2}{c|}{Merge} &2.81&36.78 &50.97\\
        \hline
        \multicolumn{2}{c|}{Post-train} &\textbf{21.11}&\textbf{45.85} &52.59\\
        \bottomrule
    \end{tabular}
    \caption{Results of single fine-tuned and merged models in ImageNet-LT with MiSLAS. }
    \label{tab:supp-single-ft-merge-imagenet-mislas}
\end{table*}

\begin{table*}[t]
    \centering
	\small
    \begin{tabular}{c|c|ccc}
        \toprule
        \multicolumn{2}{c|}{Methods} &Harmonic Mean&Geometric Mean &Accuracy\\
        \hline
        \multirow{2}{*}{Original Model} & Stage1 &1.19&39.50&66.87\\
                                & Stage2 &2.03&51.27&\textbf{71.57}\\
        \hline
        \multirow{10}{*}{Balanced Training} & Model1  &1.99&50.22&70.67\\
                                            & Model2  &1.90&49.55&70.59\\
                                            & Model3  &1.96&49.91&70.56\\
                                            & Model4  &1.96&49.87&70.50\\
                                            & Model5  &1.97&50.10&70.70\\
                                            & Model6  &1.97&49.96&70.58\\
                                            & Model7  &1.93&49.69&70.53\\
                                            & Model8  &1.93&49.89&70.72\\
                                            & Model9  &1.98&50.31&70.91\\
                                            & Model10  &1.97&49.97&70.52\\
        \hline
        \multicolumn{2}{c|}{Merge} &2.02&50.69 &70.96\\
        \hline
        \multicolumn{2}{c|}{Post-train} &\textbf{2.18}&\textbf{52.02} &71.43\\
        \bottomrule
    \end{tabular}
    \caption{Results of single fine-tuned and merged models in iNaturalist2018 with MiSLAS. }
    \label{tab:supp-single-ft-merge-ina2018-mislas}
\end{table*}

\begin{table*}[t]
    \centering
	\small
    \begin{tabular}{c|c|ccc}
        \toprule
        \multicolumn{2}{c|}{Methods} &Harmonic Mean&Geometric Mean &Accuracy\\
        \hline
        \multirow{2}{*}{Original Model} & Pre-train &21.75&51.29&\textbf{58.32}\\
                                & FC &2.19&33.82&52.12\\
        \hline
        \multirow{10}{*}{Balanced Training} & Model1  &16.55 &47.34&56.03\\
                                            & Model2  &11.33&47.17&56.08\\
                                            & Model3  &10.20&46.92&55.94\\
                                            & Model4  &11.29&46.64&55.54\\
                                            & Model5  &10.21&46.69&55.73\\
                                            & Model6  &10.01&46.09&55.41\\
                                            & Model7  &12.56&46.96&55.80\\
                                            & Model8  &9.28&46.79&55.95\\
                                            & Model9  &7.87&46.21&55.69\\
                                            & Model10  &12.65&47.33&56.02\\
        \hline
        \multicolumn{2}{c|}{Merge} &11.52&48.46 &57.03\\
        \hline
        \multicolumn{2}{c|}{Post-train} &\textbf{22.85}&\textbf{51.45} &58.15\\
        \bottomrule
    \end{tabular}
    \caption{Results of single fine-tuned and merged models in ImageNet-LT with PaCo. }
    \label{tab:supp-single-ft-merge-imagenet-paco}
\end{table*}

\begin{table*}[t]
    \centering
	\small
    \begin{tabular}{c|c|ccc}
        \toprule
        \multicolumn{2}{c|}{Methods} &Harmonic Mean&Geometric Mean &Accuracy\\
        \hline
        \multirow{2}{*}{Original Model} & Pre-train &10.93&35.00&\textbf{41.68}\\
                                & FC &1.86&16.58&30.24\\
        \hline
        \multirow{10}{*}{Balanced Training} & Model1  &10.06&31.10&38.03\\
                                            & Model2  &10.24&31.94&38.64\\
                                            & Model3  &10.16&31.64&38.43\\
                                            & Model4  &20.49&32.02&38.52\\
                                            & Model5  &13.89&32.18&38.57\\
                                            & Model6  &13.80&32.31&38.85\\
                                            & Model7  &8.26&31.40&38.24\\
                                            & Model8  &8.18&31.77&38.73\\
                                            & Model9  &10.50&31.97&38.40\\
                                            & Model10  &14.03&31.74&38.07\\
        \hline
        \multicolumn{2}{c|}{Merge} &10.61&32.85&39.43\\
        \hline
        \multicolumn{2}{c|}{Post-train} &\textbf{29.36}&\textbf{35.86} &40.72\\
        \bottomrule
    \end{tabular}
    \caption{Results of single fine-tuned and merged models in Places-LT with GPaCo. }
    \label{tab:supp-single-ft-merge-placeslt-gpaco}
\end{table*}

\section{Visualization of the Per-Class Accuracy}

\begin{figure*}[t]
   \centering
   \subcaptionbox{\label{fig:per-class-acc-stage1-merged}}{\includegraphics[width=0.48\linewidth]{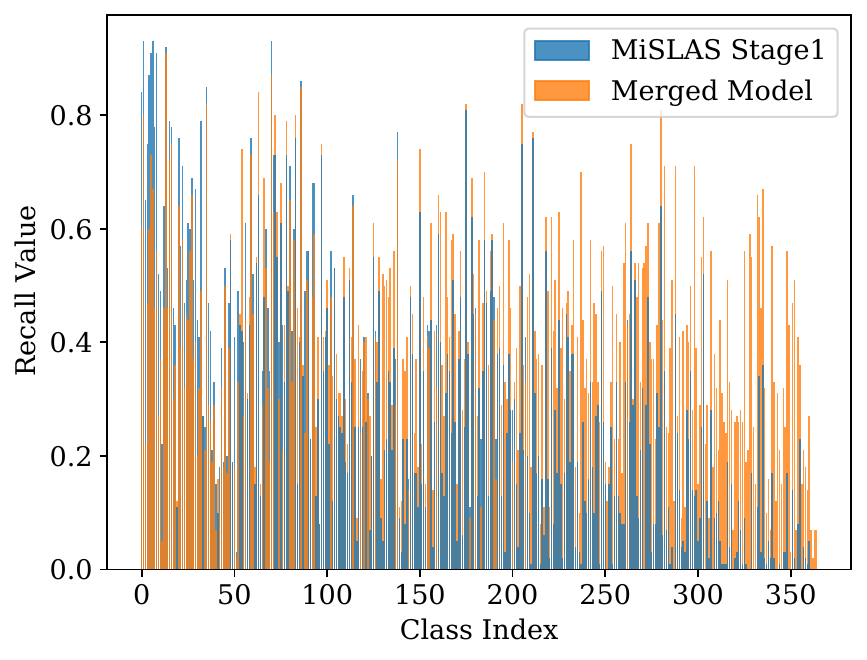}}
    \,
    \subcaptionbox{\label{fig:per-class-acc-stage2-final}}{\includegraphics[width=0.48\linewidth]{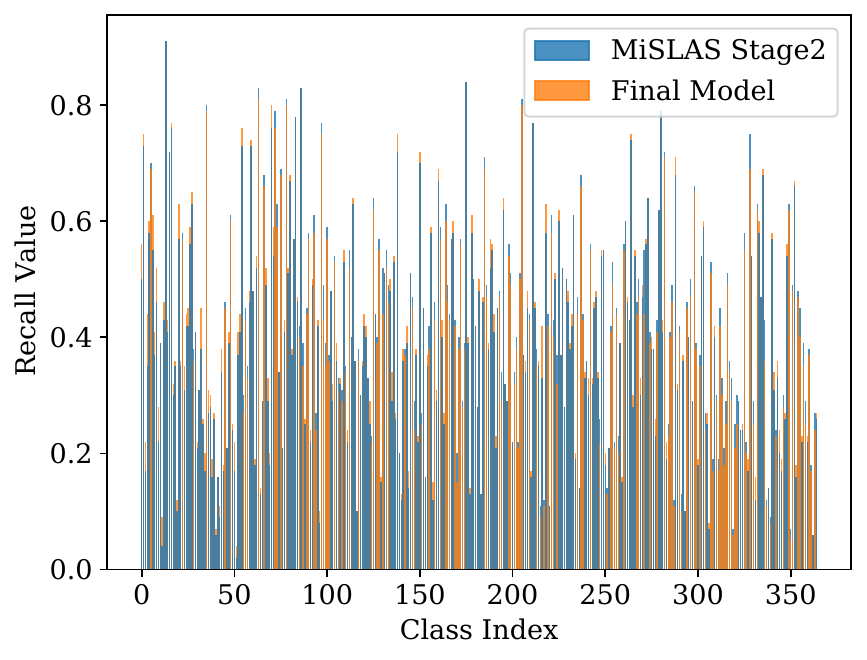}}
  \caption{Visualization of the change in the distribution of per-class recall (i.e., accuracy). (\protect\subref{fig:per-class-acc-stage1-merged}) shows that by performing balanced training on our sampled few-shot datasets and later merging all models together, we are able to greatly improve the performance of the model. (\protect\subref{fig:per-class-acc-stage2-final}) is the comparison of per-class accuracy between our final model and MiSLAS.}
  \label{fig:visualization-per-class-acc} 
\end{figure*}

Since our goal is to improve the performance of worst categories, here we visualize the change of per-class accuracy after applying our method to MiSLAS, and the result is shown in Figure~\ref{fig:visualization-per-class-acc}. As we can see from these two figures, our proposed balanced training makes the distribution of per-class accuracy more uniform, thus improves the worst categories and leads to a higher harmonic mean.

\end{document}